\documentclass{llncs}

\usepackage{amsmath, amssymb, mathtools}

\usepackage{hyperref}
\usepackage{cite}
\usepackage{enumitem}
\usepackage[ruled]{algorithm2e}
\usepackage{graphicx}
\graphicspath{{./Figs/}}

\usepackage{siunitx}
\usepackage{booktabs}

\usepackage{multirow}
\usepackage{array}

\usepackage{comment}
\usepackage{caption}
\usepackage{subcaption}
\usepackage[toc,title]{appendix}
\usepackage{epigraph}
\usepackage{bm}
\usepackage{tikz}
\usepackage{dirtytalk}
\usepackage{rotating}

\usepackage{fancyhdr}
\begin{document}
	
\title{One Framework to Rule Them All: Unifying RL-Based and RL-Free Methods in RLHF}

\author{Xin Cai}
\institute{Independent Researcher\\
	\email{xincai00@gmail.com}\hspace{2em}\href{https://totalvariation.github.io/}{Personal Webpage}}

\maketitle

\begin{abstract}
In this article, we primarily examine a variety of RL-based and RL-free methods designed to address Reinforcement Learning from Human Feedback (RLHF) and Large Reasoning Models (LRMs). We begin with a concise overview of the typical steps involved in RLHF and LRMs. Next, we reinterpret several RL-based and RL-free algorithms through the perspective of neural structured bandit prediction, providing a clear conceptual framework that uncovers a deeper connection between these seemingly distinct approaches. Following this, we briefly review some core principles of reinforcement learning, drawing attention to an often-overlooked aspect in existing RLHF studies. This leads to a detailed derivation of the standard RLHF objective within a full RL context, demonstrating its equivalence to neural structured bandit prediction. Finally, by reinvestigating the principles behind Proximal Policy Optimization (PPO), we pinpoint areas needing adjustment, which culminates in the introduction of the Generalized Reinforce Optimization (GRO) framework, seamlessly integrating RL-based and RL-free methods in RLHF.  We look forward to the community’s efforts to empirically validate GRO and invite constructive feedback.
\end{abstract}
\epigraph{Everything has already been done, every story has been told, every scene has been shot. It’s our job to do it one better.}{\textit{Stanley Kubrick}}
\section{Overview on Reinforcement Learning From Human Feedback (RLHF) and Large Reasoning Models(LRMs)}
\label{sec: overview-on-rlhf}

The pipeline of reinforcement learning from human feedback (RLHF) \cite{ziegler1909fine} comprises three stages:

\begin{itemize}
\item \textbf{Supervised Fine-Tuning (SFT)}: A large language model (LLM) pre-trained on the internet-scale corpora with next-token prediction loss is further fine-tuned using cross-entropy loss on a comparatively smaller dataset consisting of prompt-answer pairs (i.e., instruction-tuning dataset) and characterized by high-quality of responses, resulting in $\pi_{\text{ref}}$.

\item \textbf{Reward Modelling}: A human preference dataset $\mathcal{D} = \{(\mathbf{x}, \mathbf{y}^+, \mathbf{y}^-)\}_{i=1}^N$, where $\mathbf{x}$ denotes input prompts, $\mathbf{y}^+$ denotes favourable answers, and $\mathbf{y}^-$ denotes unfavourable answers, is first collected by having human annotators to rank different responses generated by $\pi_{\text{ref}}$. Then, a reward model $r_{\theta}(\cdot)$ is trained by minimizing the following objective,

\begin{equation}
\mathcal{L}_{\text{RM}} = -\mathbb{E}_{(\mathbf{x}, \mathbf{y}^+, \mathbf{y}^-)\sim \mathcal{D}}\left[\log\sigma(r_{\theta}(\mathbf{x}, \mathbf{y}^+) - r_{\theta}(\mathbf{x}, \mathbf{y}^-)) \right]\label{eq: 1-1}
\end{equation}

\noindent where $\sigma$ denotes the logistic function.

\item \textbf{RL Fine-Tuning}: In this stage, the model is optimized with the following objective through reinforcement learning, typically using Proximal Policy Optimization (PPO) \cite{schulman2017proximal},

\begin{equation}
\max_{\theta}\mathbb{E}_{\mathbf{x}\sim\mathcal{D}, \mathbf{y}\sim\pi_{\theta}(\mathbf{y}\,|\,\mathbf{x})} \left[r_{\theta}(\mathbf{x}, \mathbf{y}) - \beta D_{\text{KL}}(\pi_{\theta}(\mathbf{y}\,|\,\mathbf{x})\,\|\,\pi_{\text{ref}}(\mathbf{y}\,|\,\mathbf{x})) \right]\label{eq: 1-2}
\end{equation}

\noindent where the reward model $r_{\theta}(\cdot)$ trained previously is used to provide numerical scores on generated answers, and a KL-regularized term is introduced to prevent $\pi_{\theta}$ from concentrating probability mass on a few highest-reward responses, i.e., by keeping $\pi_{\theta}$ close to $\pi_{\text{ref}}$ in order to maintain generation diversity. 
\end{itemize}

Concerning recent advances in Large Reasoning Models (LRMs) \cite{xie2025logic, shao2024deepseekmath}, there have been two marked differences compared to RLHF. First, skipping the SFT stage has proven to produce negligible negative impact on the final performance. Second, rule-based reward functions have been demonstrated more effective than (process) reward models.  Key technical contributions in developing RL methods for optimizing the core objective \eqref{eq: 1-2} can be applied to each scenario indiscriminately. As this article is focused on understanding the algorithmic innovations in tackling the optimization problem in RLHF or LRMs, we hereafter will not particularly distinguish these two terms by omitting irrelevant details.
\section{Neural Bandit Structured Prediction}
\label{sec: bandits}

RLHF has been studied in interactive neural machine translation \cite{nguyen2017reinforcement, lam2018reinforcement, kreutzer2017bandit} or QA systems \cite{gao2022simulating} before the advent of LLMs. It is intriguing to note that those earlier works, aiming to improve the quality of a machine translation system on-the-fly with noisy and sparse user feedback, have approached the problem via the perspective of bandit structured prediction, an extension of contextual bandits, where the model predicts a structured output (e.g., a full translated sentence) as a single action in response to a given input. The problem formulation is based on the observation that if actions selected had no effect on next-state transition, then a standard RL problem would be reduced to contextual bandits where the agent observes the context, chooses an action and receives a reward only. The core challenge of contextual bandits still remains to be balancing exploration and exploitation to achieve maximum rewards, with increased complexity in language modelling tasks where  an enormous action space (i.e., language vocabulary)  and possibly delayed rewards need to be successfully handled.
 
The objective function of neural bandit structured prediction is usually formulated as follows:

\begin{align}
	\mathcal{J}(\pi_{\theta}) &= \mathbb{E}_{p(\mathbf{x})\sim \mathcal{D}, \pi_{\theta}(\mathbf{y}\,|\,\mathbf{x})}[R(\mathbf{y})]\label{eq: 2-1}\\
	\nabla_{\theta} \mathcal{J}(\pi_{\theta}) &= \mathbb{E}_{p(\mathbf{x})\sim \mathcal{D}}\left[ \sum \nabla_{\theta} \pi_{\theta}(\mathbf{y}\,|\,\mathbf{x})R(\mathbf{y}) \right]\nonumber \\
	&= \mathbb{E}_{p(\mathbf{x})\sim \mathcal{D}, \mathbf{y}\sim \pi_{\theta}(\mathbf{y}\,|\,\mathbf{x})}[\nabla_{\theta} \log(\pi_{\theta}(\mathbf{y}\,|\,\mathbf{x}))  R(\mathbf{y})]\label{eq: 2-2}
\end{align}

\noindent where $\mathbf{x} = (x_1, x_2, \dots, x_{T_{max}})$ denotes input tokens, corresponding to source sentences in machine translation or user prompts in LLMs, and $\mathbf{y} = (y_1, y_2, \dots, y_{T_{max}})$ are output tokens generated by $\pi_{\theta}(\mathbf{y}\,|\,\mathbf{x}) = \prod_{t=1}^{{T_{max}}} \pi_{\theta}(y_t \,|\, \mathbf{y}_{<t}, \mathbf{x})$ in an autoregressive fashion.  Assuming autoregressive modelling and sequence-level\footnote{We deliberately choose not to use the term "trajectory" to describe full completions generated by LLMs in order to avoid potential confusion with its counterpart concept in RL.} rewards, the Monte Carlo Estimate of the gradient (i.e., REINFORCE) can be further simplified as follows.

\begin{align}
	\hat{g} = \frac{1}{N}\sum_{i=1}^N \sum_{t=1}^{T_{max}}R(\mathbf{y}^i)\nabla_{\theta}\log\pi_{\theta}(y_t^i\,|\,\mathbf{y}_{<t}^i, \mathbf{x})\; \text{where}\: \mathbf{y}^1, \dots, \mathbf{y}^N \overset{\makebox[0pt]{\mbox{\normalfont\tiny\sffamily i.i.d}}}{\sim} \pi_{\theta}(\cdot\,|\,\mathbf{x})\label{eq: 2-3}
\end{align}

It is well-known that the REINFORCE gradient estimator suffers from high-variance, which can be mitigated by introducing control variates or baselines, leading to gradient estimators underpinning existing RLHF works as shown in \eqref{eq: 2-4}.

\begin{align}
	\hat{g} &= \frac{1}{N}\sum_{i=1}^N \sum_{t=1}^{T_{max}}\left(R(\mathbf{y}^i) - B\right)\nabla_{\theta}\log\pi_{\theta}(y_t^i\,|\,\mathbf{y}_{<t}^i, \mathbf{x})\; \text{where}\: \mathbf{y}^1, \dots, \mathbf{y}^N \overset{\makebox[0pt]{\mbox{\normalfont\tiny\sffamily i.i.d}}}{\sim} \pi_{\theta}(\cdot\,|\,\mathbf{x})\label{eq: 2-4}
\end{align}

\noindent where as long as $B$ is independent to $y_t$, the derived gradient estimator is unbiased shown in Eq.\eqref{eq: 2-5}.

\begin{align}
	\mathbb{E}_{y_t \sim \pi_{\theta}(y_t\,|\,\mathbf{y}_{<t}, \mathbf{x})}\left[B \nabla_{\theta}\log(\pi_{\theta}(y_t\,|\,\mathbf{y}_{<t}, \mathbf{x}))  \right] &= \sum_{y_t}  \nabla_{\theta} \pi_{\theta}(y_t\,|\,\mathbf{y}_{<t}, \mathbf{x}) B \nonumber \\
	&= B \nabla_{\theta} \sum_{y_t} \pi_{\theta}(y_t\,|\,\mathbf{y}_{<t}, \mathbf{x}) \nonumber \\
	&= B \nabla_{\theta} \sum_{y_t} 1 = 0\label{eq: 2-5}
\end{align}

A general pipeline of neural bandit structured prediction is shown in Algorithm \ref{algo: 2-1}. It is worth noting that there is no marked algorithmic difference between building interactive neural machine translation models and employing RLHF to refine LLMs.

\IncMargin{1em}
\begin{algorithm}[H]
\textbf{Goal}: Learn a policy network $\pi_{\theta}$ by maximizing $\mathcal{J}(\pi_{\theta})$ with user or simulated feedback.\\
\KwIn {Dataset $D$, Task Evaluation Criterion or Reward Model $R$, A policy model initialized with $\pi_{\theta_0}$}
\BlankLine

\ForEach{$k = 0, 1, 2, \dots, \text{max\_iter}$}{
	Draw $\mathbf{x}_k$ from the dataset $D$\;
	Sample $\mathbf{y}_k \sim \pi_{\theta}(\mathbf{y}\,|\,\mathbf{x}_k)$\;
	Receive feedback $R(\mathbf{y}_k)$\;
	$\pi_{\theta_{k+1}} = \pi_{\theta_k} + \alpha \hat{g}$\;
}
\caption{Neural Bandit Structured Prediction}\label{algo: 2-1}
\end{algorithm}\DecMargin{1em}

To gain more intuition concerning the inner workings of REINFORCE, we can take a closer look at the gradient estimator, rewritten as follows for the illustrative purpose.

\begin{align}
	\pi_{\theta_{t+1}} &= \pi_{\theta_t} + \alpha (R(\mathbf{y}) - B) \nabla_{\theta}\log(\pi_{\theta_t}(y_t\,|\,\mathbf{y}_{<t}, \mathbf{x}))\nonumber \\
	&= \pi_{\theta_t} + \alpha \underbrace{\left(\frac{R(\mathbf{y}) - B}{\pi_{\theta_t}(y_t\,|\,\mathbf{y}_{<t}, \mathbf{x})}\right)}_{\beta_t} \nabla_{\theta}\pi_{\theta_t}(y_t\,|\,\mathbf{y}_{<t}, \mathbf{x})\label{eq: 2-6}
\end{align}

\begin{align}
	\pi_{\theta_{t+1}} &= \pi_{\theta_t} + \left(\nabla_{\theta}\pi_{\theta_t}(y_t\,|\,\mathbf{y}_{<t}, \mathbf{x}) \right)^T(\pi_{\theta_{t+1}} - \pi_{\theta_t})\nonumber \\
	&= \pi_{\theta_t} + \alpha \beta_t \|\nabla_{\theta}\pi_{\theta_t}(y_t\,|\,\mathbf{y}_{<t}, \mathbf{x}) \|_2^2 \label{eq: 2-7}
\end{align}

Based on the first-order Taylor expansion shown in Eq.\eqref{eq: 2-7}, it can be seen that the probability of choosing an action $\pi_{\theta_{t+1}}(y_t\,|\,\mathbf{y}_{<t}, \mathbf{x})$ is increased provided $\beta_t > 0$, i.e., actions leading to higher rewards compared to the baseline are encouraged, and vice versa. More crucially, $\beta_t$ is inversely proportional to $\pi_{\theta}(y_t\,|\,\mathbf{y}_{<t}, \mathbf{x})$ when $R(\mathbf{y}) - B > 0$, serving as the key driver to incentivize exploration for searching better candidate solutions in contrast to supervised learning. Therefore, it is not surprising to see that learning conducted in the pure form of instructive feedback (supervised learning) results in memorization, whereas machines harnessing the potential of active exploration with evaluative feedback can generalize better \cite{xie2025logic}.

\subsection{RL-Based Methods in RLHF}
\label{subsec: rl-based}

Facilitated by such simplifications, we can easily summarize recent representative works in RLHF or LRMs (RLOO, \cite{ahmadian2024back}, GRPO \cite{shao2024deepseekmath}, ReMax \cite{li2023remax}, REINFORCE++ \cite{hu2025reinforce++}) in the following.

\begin{align}
 \text{RLOO} &: \frac{1}{N}\sum_{i=1}^N\sum_{t=1}^{T_{\text{max}}}\left(R(\mathbf{y}^i) - \frac{1}{N-1}\sum_{j\neq i}R(\mathbf{y}^j)\right) \nabla_{\theta}\log\pi_{\theta}(y_t^i\,|\,\mathbf{y}_{<t}^i, \mathbf{x}) \label{eq: 2-8}\\
\text{GRPO}&: \frac{1}{N}\sum_{i=1}^N \sum_{t=1}^{T_{\text{max}}}\left(\frac{R(\mathbf{y}^i) - \mu_g}{\sigma_g}\right)\nabla_{\theta}\log\pi_{\theta}(y_t^i\,|\,\mathbf{y}_{<t}^i, \mathbf{x})\label{eq: 2-9}\\
&\quad\text{where}\: \mu_g = \sum_{i=1}^N R(\mathbf{y}^i), \sigma_g = \sqrt{\frac{\sum_{i=1}^N (R(\mathbf{y}^i) - \mu_g)^2}{N}}\nonumber \\
\text{ReMax}&:\frac{1}{N}\sum_{i=1}^N \sum_{t=1}^{T_{\text{max}}}\left(R(\mathbf{y}^i) - R(\mathbf{y}^{\text{greedy}})\right)\nabla_{\theta}\log\pi_{\theta}(y_t^i\,|\,\mathbf{y}_{<t}^i, \mathbf{x}) \label{eq: 2-10}\\
&\quad\text{where}\: y_t^{\text{greedy}} \sim \mathrm{argmax}_{y_t}\pi_{\theta}(\cdot\,|\,\mathbf{y}_{<t}^{\text{greedy}}, \mathbf{x})\nonumber \\
\text{REINFORCE++}&: \frac{1}{N}\sum_{i=1}^N \sum_{t=1}^{T_{\text{max}}} \hat{R}(y_t^i) \left(\frac{\pi_{\theta}(y_t^i\,|\,\mathbf{y}_{<t}^i, \mathbf{x})}{\pi_{\theta_{\text{old}}}(y_t^i\,|\,\mathbf{y}_{<t}^i, \mathbf{x})}\right)\nabla_{\theta}\log\pi_{\theta}(y_t^i\,|\,\mathbf{y}_{<t}^i, \mathbf{x}) \label{eq: 2-11} \\
&\quad\text{where}\: \hat{R}(y_t^i) = \sum_{k=t}^{T_{\text{max}}} \gamma^{k-t} \log\frac{\pi_{\theta}(y_t\,|\,\mathbf{y}_{<t}^i, \mathbf{x})}{\pi_{\text{ref}(y_t\,|\,\mathbf{y}_{<t}^i, \mathbf{x})}} + \gamma^{T_{\text{max}}-t}(R(\mathbf{y}^i) - \mu_g), \nonumber\\
&\quad\quad\mu_g=\sum_{i=1}^N R(\mathbf{y}^i), \,\gamma \in (0, 1) \nonumber
\end{align}

RLOO \cite{ahmadian2024back}, GRPO \cite{shao2024deepseekmath}, and ReMax \cite{li2023remax} use rule-based reward functions evaluating on full completions with slight differences in the baseline design. REINFORCE++ \cite{hu2025reinforce++}, however, adopts a token-level reward strategy where the per-token KL constraint is employed as a form of knowledge distillation (i.e., a simple way to introduce process reward models (PRMs)) in addition to the sequence-level reward obtained by fulfilling specified rules. Another distinction in REINFORCE++ \cite{hu2025reinforce++} is the added importance weight $r_t(\theta)=\frac{\pi_{\theta}(y_t^i\,|\,\mathbf{y}_{<t}^i, \mathbf{x})}{\pi_{\theta_{\text{old}}}(y_t^i\,|\,\mathbf{y}_{<t}^i, \mathbf{x})}$ inherited from PPO \cite{schulman2017proximal}, allowing multiple updates of model parameters on the freshly collected samples with $\pi_{\theta_{\text{old}}}$ to maintain unbiased estimates. Please note that for clarity we omit cases where gradient is zero in REINFORCE++ \cite{hu2025reinforce++} caused by the clipping or min operator, intended to prevent excessive optimization on already correctly made decisions such that the probability ratio has fallen outside the clipping interval $(1 - \epsilon, 1 + \epsilon)$.

\subsection{RL-Free Methods in RLHF}
\label{subsec: rl-free}

It seems a bit of counter intuitive to formulate \say{RL-free} methods via the lens of bandit structured prediction. But with a dissection of gradients of  \say{RL-free} methods, we can observe the underlying close connection.

The following is the objective function and respective gradient of Direct Preference Optimization (DPO) \cite{rafailov2023direct}.

\begin{align}
\mathcal{L}_{\text{DPO}}(\pi_{\theta};\pi_{\text{ref}}) &= \mathbb{E}_{(\mathbf{x}, \mathbf{y}^+, \mathbf{y}^-)\sim \mathcal{D}_{\text{pref}}}\left[\log\sigma\left(\beta\log\frac{\pi_{\theta}(\mathbf{y}^+\,|\,\mathbf{x})}{\pi_{\text{ref}}(\mathbf{y}^+\,|\,\mathbf{x})} - \beta\log\frac{\pi_{\theta}(\mathbf{y}^-\,|\,\mathbf{x})}{\pi_{\text{ref}}(\mathbf{y}^-\,|\,\mathbf{x})} \right) \right] \label{eq: 2-12} \\
\nabla_{\theta}\mathcal{L}_{\text{DPO}}(\pi_{\theta};\pi_{\text{ref}}) &= \beta \mathbb{E}_{(\mathbf{x}, \mathbf{y}^+, \mathbf{y}^-)\sim \mathcal{D}_{\text{pref}}}\left[\sigma\left(\hat{r}_{\theta}(\mathbf{x}, \mathbf{y}^-) - \hat{r}_{\theta}(\mathbf{x}, \mathbf{y}^+)\right)\left[\nabla_{\theta}\log\pi(\mathbf{y}^+\,|\,\mathbf{x}) -  \nabla_{\theta}\log\pi(\mathbf{y}^-\,|\,\mathbf{x})\right] \right] \label{eq: 2-13}\\
&\quad\text{where}\: \hat{r}_{\theta}(\mathbf{x}, \mathbf{y}) = \beta\log\frac{\pi_{\theta}(\mathbf{y}\,|\,\mathbf{x})}{\pi_{\text{ref}}(\mathbf{y}\,|\,\mathbf{x})} = \beta\sum_t \log\frac{\pi_{\theta}(y_t\,|\,\mathbf{y}_{<t}, \mathbf{x})}{\pi_{\text{ref}}(y_t\,|\,\mathbf{y}_{<t}, \mathbf{x})}\nonumber
\end{align}

With minor tweaks, DPO \cite{rafailov2023direct} can be derived as a simple variant of REINFORCE gradient estimator, as illustrated in the following:

\begin{align}
\nabla_{\theta}\mathcal{L}_{\text{DPO}}(\pi_{\theta};\pi_{\text{ref}}) &= \beta\sigma\left(\hat{r}_{\theta}(\mathbf{x}, \mathbf{y}^-) - \hat{r}_{\theta}(\mathbf{x}, \mathbf{y}^+)\right) \frac{1}{2}\sum_{i=1}^2 \left[(R(\mathbf{y}^+) - B) \nabla_{\theta}\log\pi_{\theta}(\mathbf{y}^+\,|\,\mathbf{x}) \right. \nonumber\\
&\quad \left. + (R(\mathbf{y}^-) - B)\nabla_{\theta}\log\pi_{\theta}(\mathbf{y}^-\,|\,\mathbf{x})\right]\nonumber \\
&= \beta\sigma\left(\hat{r}_{\theta}(\mathbf{x}, \mathbf{y}^-) - \hat{r}_{\theta}(\mathbf{x}, \mathbf{y}^+)\right) \frac{1}{2}\sum_{i=1}^2 \left[(R(\mathbf{y}^+) - B)\sum_t \nabla_{\theta}\log\pi_{\theta}(y^+_t\,|\,\mathbf{y}^+_{<t},\mathbf{x}) \right. \nonumber\\ 
&\quad \left. + (R(\mathbf{y}^-) - B)\sum_t \nabla_{\theta}\log\pi_{\theta}(y^-_t\,|\,\mathbf{y}^-_{<t},\mathbf{x}) \right] \label{eq: 2-14} \\
&\quad\text{where}\: R(\mathbf{y}^+) = 1, R(\mathbf{y}^-) = -1, B=\frac{R(\mathbf{y}^+) + R(\mathbf{y}^-)}{2} = 0 \nonumber
\end{align}

The only difference between the DPO \cite{rafailov2023direct} gradient estimator and the REINFORCE is the added weighting factor $\sigma\left(\hat{r}_{\theta}(\mathbf{x}, \mathbf{y}^-) - \hat{r}_{\theta}(\mathbf{x}, \mathbf{y}^+)\right)$, assigning higher weights to sample pairs in the degree to which they have been misordered, which can be viewed as introducing the concept of margin into the learning procedure and thus implicitly imposing hard positive/negative mining. Aided by the perspective of understanding DPO \cite{rafailov2023direct} by dissecting its gradient, it is clear to spot that the performance of DPO \cite{rafailov2023direct} is restricted by the small sample size (2) for Monte Carlo estimates and the inflexibility to accommodate diverse reward functions.

We then quickly examine the gradient of KTO \cite{ethayarajh2024kto} in the following, which can be viewed as an extension of DPO \cite{rafailov2023direct} framed through the lens of Kahneman \& Tversky’s prospect theory \cite{tversky1992advances}.

\begin{align}
\nabla_{\theta}\mathcal{L}_{\text{KTO}}(\pi_{\theta};\pi_{\text{ref}}) &= \beta\lambda_{\mathbf{y}^i}(R(\mathbf{y}^i) - B)\sum_t \left[ \sigma\left(\beta(r_{\theta}(\mathbf{x}, y^i_t) - A) \right)\left(1 -\sigma\left(\beta(r_{\theta}(\mathbf{x}, y^i_t) - A) \right) \right) \right. \nonumber\\
&\quad\left. \nabla_{\theta} \log\pi_{\theta}(y^i_t\,|\,\mathbf{y}^i_{<t},\mathbf{x})\right] \label{eq: 2-15} \\
&\quad \text{where}\: R(\mathbf{y}^+) = 1, R(\mathbf{y}^-) = -1, B=\frac{R(\mathbf{y}^+) + R(\mathbf{y}^-)}{2} = 0\nonumber \\
&\quad r_{\theta}(\mathbf{x}, y^i_t) = \log\frac{\pi_{\theta}(y^i_t\,|\,\mathbf{y}^i_{<t},\mathbf{x})}{\pi_{\text{ref}}(y^i_t\,|\,\mathbf{y}^i_{<t},\mathbf{x})}, \nonumber \\
&\quad A = \max(0, \frac{1}{m}\sum_i \sum_t \log\frac{\pi_{\theta}(y^j_t\,|\,\mathbf{y}^{j}_{<t},\mathbf{x}^i)}{\pi_{\text{ref}}(y^j_t\,|\,\mathbf{y}^{j}_{<t},\mathbf{x}^i)}), j = (i + 1) \operatorname{mod} m, \nonumber \\
&\quad \lambda_{\mathbf{y}^i}=
		\begin{cases}  
			\lambda_D, & \mathbf{y}^i=\mathbf{y}^+\:\text{desirable},\\
			\lambda_U, & \mathbf{y}^i=\mathbf{y}^- \:\text{undesirable}
		\end{cases}\nonumber
\end{align}

It is worth noting that in the original works of DPO \cite{rafailov2023direct} and KTO \cite{ethayarajh2024kto},  the log ratio $r_{\theta}(\mathbf{x}, y^i_t)$ has been interpreted as reward signals. We argue that the true reward is $+1$ for positive responses and $-1$ for negative responses. Imagine if we flipped the sign of positive and negative responses, the model would be optimized towards the exact opposite direction, favouring negative responses than positive responses. Therefore, we argue that either $\sigma\left(\hat{r}_{\theta}(\mathbf{x}, \mathbf{y}^-) - \hat{r}_{\theta}(\mathbf{x}, \mathbf{y}^+)\right)$ in DPO \cite{rafailov2023direct} or $\sigma\left(\beta(r_{\theta}(\mathbf{x}, y^i_t) - A) \right)\left(1 -\sigma\left(\beta(r_{\theta}(\mathbf{x}, y^i_t) - A) \right) \right)$ in KTO \cite{ethayarajh2024kto} is more appropriate to be viewed as dynamic weighting factors, preventing excessive optimization over already well-distinguished samples and thus acting as a soft version of the clipping and min operator in PPO \cite{schulman2017proximal}.

Subsequently, we give a cursory glance to the training objective of Contrastive Preference Learning (CPL) \cite{hejna2023contrastive} and its gradient.

\begin{align}
\mathcal{L}_{\text{CPL}}(\pi_{\theta}, \mathcal{D}_{\text{pref}}) &= \mathbb{E}_{(\mathbf{x}, \mathbf{y}^+, \mathbf{y}^-)\sim \mathcal{D}_{\text{pref}}}\left[\right.\nonumber\\
&\quad \left. \log\frac{\exp\sum_t \gamma^t\beta\log\pi_{\theta}(y^+_t\,|\,\mathbf{y}^+_{<t}, \mathbf{x})}{\exp\sum_t \gamma^t\beta\log\pi_{\theta}(y^+_t\,|\,\mathbf{y}^+_{<t},\mathbf{x}) + \exp\sum_t \gamma^t\beta\log\pi_{\theta}(y^-_t\,|\,\mathbf{y}^-_{<t},\mathbf{x})} \right] \label{eq: 2-16} \\
\nabla_{\theta}\mathcal{L}_{\text{CPL}}(\pi_{\theta}, \mathcal{D}_{\text{pref}}) &= \beta \mathbb{E}_{(\mathbf{x}, \mathbf{y}^+, \mathbf{y}^-)\sim \mathcal{D}_{\text{pref}}}\left[\sigma\left(\hat{r}_{\theta}(\mathbf{x}, \mathbf{y}^-) - \hat{r}_{\theta}(\mathbf{x}, \mathbf{y}^+)\right) \right.\nonumber\\
&\quad \left. \left[\sum_t\gamma^t\nabla_{\theta}\log\pi_{\theta}(y^+_t\,|\,\mathbf{y}^+_{<t},\mathbf{x}) -  \sum_t\gamma^t\nabla_{\theta}\log\pi_{\theta}(y^-_t\,|\,\mathbf{y}^-_{<t},\mathbf{x})\right] \right]\label{eq: 2-17}\\
&\quad\text{where}\: \hat{r}_{\theta}(\mathbf{x}, \mathbf{y}) = \beta\sum_t\gamma^t\log\pi_{\theta}(y_t\,|\,\mathbf{y}_{<t},\mathbf{x}) \nonumber
\end{align}

With the same modifications in the derivation of DPO \cite{rafailov2023direct} gradient estimator, it can be seen that the CPL \cite{hejna2023contrastive} gradient bears a close resemblance with that of DPO \cite{rafailov2023direct} although two works have been presented in a rather different way.

\begin{align}
\nabla_{\theta}\mathcal{L}_{\text{CPL}}(\pi_{\theta}, \mathcal{D}_{\text{pref}}) &= \beta\sigma\left(\hat{r}_{\theta}(\mathbf{x}, \mathbf{y}^-) - \hat{r}_{\theta}(\mathbf{x}, \mathbf{y}^+)\right) \frac{1}{2}\sum_{i=1}^2 \left[(R(\mathbf{y}^+) - B)\sum_t \gamma^t \nabla_{\theta}\log\pi(y^+_t\,|\,\mathbf{y}^+_{<t},\mathbf{x}) \right.\nonumber\\
&\quad \left. + (R(\mathbf{y}^-) - B)\sum_t \gamma^t \nabla_{\theta}\log\pi(y^-_t\,|\,\mathbf{y}^-_{<t},\mathbf{x}) \right] \label{eq: 2-18} \\
&\quad\text{where}\: R(\mathbf{y}^+) = 1, R(\mathbf{y}^-) = -1, B=\frac{R(\mathbf{y}^+) + R(\mathbf{y}^-)}{2} = 0 \nonumber
\end{align}

By inspecting the objective functions in DPO \cite{rafailov2023direct} and CPL \cite{hejna2023contrastive}, if $\pi_{\text{ref}}(\mathbf{y}\,|\,\mathbf{x})$ were set to the uniform prior, the objective of DPO \cite{rafailov2023direct} would be equivalent to that of CPL \cite{hejna2023contrastive} with a minor tweak in reward signals. Furthermore,  two reward functions from the same equivalence class, i.e., $r^{\prime}(\mathbf{x}, \mathbf{y}) = r(\mathbf{x}, \mathbf{y}) + f(\mathbf{x})$, induce the same optimal policy and the same preference distribution under the Bradley-Terry preference framework (Theorem 1 in DPO \cite{rafailov2023direct}), implying the fundamental equivalence between DPO \cite{rafailov2023direct} and CPL \cite{hejna2023contrastive}.

\begin{align}
\text{DPO}&: \max_{\pi_{\theta}}\mathbb{E}_{\mathbf{x}\sim\mathcal{D}, \mathbf{y}\sim\pi_{\theta}}\left[r(\mathbf{x}, \mathbf{y}) \right] - \beta\mathbb{D}_{\text{KL}}\left[\pi_{\theta}(\mathbf{y}\,|\,\mathbf{x})\,\|\,\pi_{\text{ref}}(\mathbf{y}\,|\,\mathbf{x}) \right] \label{eq: 2-19}\\
\text{CPL}&: \max_{\pi_{\theta}}\mathbb{E}_{\mathbf{x}\sim\mathcal{D}, \mathbf{y}\sim\pi_{\theta}}\left[\underbrace{r(\mathbf{x}, \mathbf{y}) - \log\sum_{y_1,\dots,y_{T_{\text{max}}}}\exp(\frac{1}{\beta}r(\mathbf{x}, \mathbf{y}))}_{A(\mathbf{x}, \mathbf{y})}\right] - \beta\mathcal{H}\left(\pi_{\theta}(\mathbf{y}\,|\,\mathbf{x}) \right) \label{eq: 2-20}
\end{align}

Yet, if we recognize that the REINFORCE gradient estimator serves as the core mechanism in RLHF, we could instead develop additional innovations directly from the Eq. \eqref{eq: 2-4}, benefiting from a clearer and simpler reasoning trace. Summarizing the recently developed variants of the REINFORCE gradient estimator in RLHF, further improvements can be made in three directions: 1) reward engineering, 2) baseline design, and 3) dynamic weighting factors where contrastive learning holds the promise of reshaping the distribution of full completions in a more desirable manner and increase sample efficiency, e.g., solutions to the same task prompt should ideally form a coherent cluster. 

\section{What could be wrong with RLHF?}
\label{sec: rl-principle}

In this section, we will be revisiting some fundamental principles of RL and attempt to diagnose what could be overlooked in formulating RLHF in the full RL framework, offering a complementary view to the previous analysis on existing RLHF methods.

Let us first review Policy Gradient Theorem \cite{zhao2024mathematical} and make some clarifications necessary. Given a Markov Decision Process (MDP), denoted by $\mathcal{M} = (\mathcal{S}, \mathcal{A}, P, r, d_0, \gamma)$, where $\mathcal{S}$ is the state space, $\mathcal{A}$ is the action space, $P$ is the transition probability defined on $\mathcal{S}\times\mathcal{A}\times\mathcal{S} \rightarrow [0, 1]$, $r$ is the reward function $\mathcal{S}\times\mathcal{A} \rightarrow \mathbb{R}$, $d_0$ is the initial state distribution, and $\gamma \in (0, 1)$ is the discounting factor, in RL literature, we usually encounter the expected discounted reward expressed as follows.

\begin{align}
\mathcal{J}(\pi) &= \mathbb{E}_{\tau \sim p_{\pi}(\tau)}\left[\sum_{t=0}^{\infty} \gamma^t r(s_t, a_t) \right] \nonumber \\
	&= \mathbb{E}_{s_0, a_0, s_1, a_1, \dots}\left[\sum_{t=0}^{\infty} \gamma^t r(s_t, a_t) \right]\label{eq: 3-1}\\
	&\quad \text{where}:\; s_0 \sim d_0, a_t \sim \pi(a_t\,|\,s_t), s_{t+1} \sim P(s_{t+1}\,|\,s_t, a_t)\nonumber
\end{align}

The cornerstone of RL is built upon the Bellman equation and Markov property, simplifying the above expectation taken over all possible trajectories under policy $\pi$. We use the following notation to denote:

\begin{itemize}
\item State value: $V^{\pi}(s_t) = \mathbb{E}_{a_t, s_{t+1}, \dots, \sim \pi(\tau\,|\,s_t)} [\sum_{k=0}^{\infty}\gamma^k r(s_{t+k}, a_{t+k})]$;
\item State-action value: $Q^{\pi}(s_t, a_t) = \mathbb{E}_{s_{t+1}, \dots, \sim \pi(\tau\,|\,s_t, a_t)} [\sum_{k=0}^{\infty}\gamma^k r(s_{t+k}, a_{t+k})]$;
\item Advantage: $A^{\pi}(s_t, a_t) = Q^{\pi}(s_t, a_t) - V^{\pi}(s_t)$;
\item Relation between state value and state-action value: $V^{\pi}(s) = \sum_a \pi(a\,|\,s) Q^{\pi}(s, a)$;
\item Bellman equation: $Q^{\pi}(s, a) = r(s, a) + \gamma \sum_{s^{\prime}, a^{\prime}}P(s^{\prime}\,|\,s, a)\pi(a^{\prime}\,|\,s^{\prime})Q^{\pi}(s^{\prime}, a^{\prime})\label{eq: 3-2}$.
\end{itemize}

Therefore, we can rewrite the objective of maximizing expected discounted rewards succinctly as $\mathcal{J}(\pi) = \mathbb{E}_{s_0}[V^{\pi}(s_0)]$ and the policy gradient theorem can be derived assisted with the Bellman equation \eqref{eq: 3-2}.

\begin{theorem}[Policy Gradient Theorem]
\label{thm: policy-gradient}
The gradient of $\mathcal{J}(\pi_{\theta}) = \mathbb{E}[\sum_{t=0}^{\infty}\gamma^t R_{t+1}]=\sum_{s\in\mathcal{S}}d_0(s)V^{\pi}(s)$ is 
\begin{align*}
\nabla_{\theta} \mathcal{J}(\pi) &= \sum_s d_0(s)\nabla_{\theta}V^{\pi}(s)\\
	&= \sum_s d_0(s) \sum_{s^{\prime}} \operatorname{Pr}_{\pi}(s^{\prime}\,|\,s)\sum_a \nabla_{\theta}\pi(a\,|\,s^{\prime})Q^{\pi}(s^{\prime}, a)\\
	&= \sum_{s^{\prime}} \left(\sum_s d_0(s)\operatorname{Pr}_{\pi}(s^{\prime}\,|\,s) \right)\sum_a \nabla_{\theta}\pi(a\,|\,s^{\prime})Q^{\pi}(s^{\prime}, a)\\
	&= \sum_{s^{\prime}} \rho_{\pi}(s^{\prime})\sum_a \nabla_{\theta}\pi(a\,|\,s^{\prime})Q^{\pi}(s^{\prime}, a)\\
	&=\sum_s \rho_{\pi}(s)\sum_a \nabla_{\theta}\pi(a\,|\,s)Q^{\pi}(s, a)\\
	&=\sum_s\rho_{\pi}(s)\sum_a \pi(a\,|\,s) \nabla_{\theta}\log\pi(a\,|\,s)Q^{\pi}(s, a)\\
	&=\mathbb{E}_{S\sim\rho_{\pi}, A\sim \pi} \left[\nabla_{\theta}\log\pi(A\,|\,S)Q^{\pi}(A, S) \right]
\end{align*}
where $d_0$ is the initial state distribution, $\rho_{\pi}(s) = \sum_{s^{\prime}\in\mathcal{S}}d_0(s^{\prime})\mathrm{Pr}_{\pi}(s\,|\,s^{\prime})$, $\mathrm{Pr}_{\pi}(s\,|\,s^{\prime})=\sum_{k=0}^{\infty}\gamma^k\left[P_{\pi}^k\right]_{s^{\prime}s}$, and $\left[P_{\pi}^k\right]_{s^{\prime}s}=\mathrm{Pr}\left( S_{t_k}=s\,|\,S_{t_1}=s{\prime} \right)$ denotes the probability of using exactly $k$ steps to transition from the state $s^{\prime}$ to $s$ under $\pi_{\theta}$. The proof of the expression of $\nabla_{\theta}V^{\pi}(s)$ is deferred to the Appendix \eqref{eq: a1-1}.
\end{theorem}

Now, it is ripe to make a comparison side by side between the gradient derived from the objective used in RLHF and that of full RL.

\begin{align*}
\text{Policy Gradient (Full RL)}:& \mathbb{E}_{S\sim\rho_{\pi}, A\sim \pi} \left[\nabla_{\theta}\log\pi(A\,|\,S)Q^{\pi}(A, S) \right]\\
\text{RLHF (Bandits)}:& \mathbb{E}_{p(\mathbf{x})\sim \mathcal{D}, \mathbf{y}\sim \pi_{\theta}(\mathbf{y}\,|\,\mathbf{x})}[\nabla_{\theta} \log(\pi_{\theta}(\mathbf{y}\,|\,\mathbf{x})) (R(\mathbf{y}) - B)]
\end{align*}

Acute readers should have noticed, in the existing literature of RLHF or RL fine-tuning of LLMs, a largely overlooked fact is, within the full RL context, states should be sampled from $\rho_{\pi}$, which represents the unnormalized discounted state distribution (i.e., occupancy measure, visitation frequency) under $\pi$, necessitating RL agents to roll out a sufficient number of steps in order to approach better estimation over $\rho_{\pi}$ (i.e., the sample efficiency challenge of on-policy RL). It is worth highlighting that actions selected in the full RL setting exert influence on the evolvement of trajectories, i.e., $P(s^{\prime}\,|\,s)=\sum_a \pi(a\,|\,s)P(s^{\prime}\,|\,s, a)$, where $P(s^{\prime}\,|\,s, a)$ is system dynamics. Therefore, in order to attain the maximum of expected cumulative rewards, $\pi_{\theta}$ needs to be optimized to choose actions leading to reward-rich states in addition to maximizing immediate rewards, reflected by the essence of Bellman equation $Q^{\pi}(s, a) = r(s, a) + \gamma \sum_{s^{\prime}}P(s^{\prime}\,|\,s, a)V^{\pi}(s^{\prime})$. In the RLHF setting, however, $p(\mathbf{x})\sim\mathcal{D}$ denotes input promts  sampled independently to each other from a static dataset, which do not continuously evolve to next states. Furthermore, the collective tokens contained in a full completion $[y_1, y_2, \dots, y_{T_{\text{max}}}]$ is more appropriate to be viewed as a single action in response to the contextual input $\mathbf{x}$, receiving afterwards an immediate reward and having no effect on the selection of the next input prompt.

If we insist on formulating the generation process of LLMs in the MDP framework, where the action is equivalent to the token predicted from the language vocabulary $a_t = y_t \in \mathcal{V}$, the state is the concatenation of the input prompt and the history of predicted tokens $s_t = [\mathbf{y}_{<t}, \mathbf{x}]$ ,  the state transition is deterministic as the next state is obtained via a concatenation $s_{t+1}= [s_t, a_t]$, and the LLM is interpreted as the policy network $\pi_{\theta}(a_t\,|\,s_t)$ with autoregressive modelling $\pi(\mathbf{a}\,|\,s_0)=\prod_{t=0}^{T_{\text{max}}}\pi(a_t\,|\,s_t)$, then the RLHF objective can be expressed ostensibly identically to the RL objective $\max_{\pi{\theta}} \mathbb{E}_{\tau\sim\rho_{\pi}(\tau)} [\sum_{t=0}^{T_{\text{max}}} r(s_t, a_t)]$\footnote{Please note there we omit the KL regularization term, which can be assimilated into per-token reward.}, which is the source of common misconception. Let us go through the gradient taking into account the deterministic state transition. \footnote{The same unrolling trick \eqref{eq: a1-1} is applied as in the proof of Policy Gradient Theorem and we set $\gamma$ to $1$.}

\begin{align}
\nabla V^{\pi}(s_0)&= \sum_{a_0}\nabla \pi(a_0\,|\,s_0) Q^{\pi}(s_0, a_0) + \sum_{s_1}P(s_1\,|\,s_0) \sum_{a_1} \nabla \pi(a_1\,|\,s_1) Q^{\pi}(s_1, a_1) \nonumber \\
&\quad + \sum_{s_1}P(s_1\,|\,s_0) \sum_{s_2}P(s_2\,|\,s_1) \nabla V^{\pi}(s_2)\nonumber \\
&= \sum_{a_0}\nabla \pi(a_0\,|\,s_0) Q^{\pi}(s_0, a_0) +  \sum_{a_1} \nabla \pi(a_1\,|\,s_1) Q^{\pi}(s_1, a_1) + \nabla V^{\pi}(s_2) \nonumber \\
&\vdots \quad\text{unrolling} \nonumber\\
&= \sum_{t=0}^{T_{\text{max}}} \sum_{a_t}\nabla \pi(a_t\,|\,s_t) Q^{\pi}(s_t, a_t) \label{eq: 3-3}
\end{align}

\begin{align}
\nabla_{\theta}\mathbb{E}_{\tau\sim\rho_{\pi}(\tau)} [\sum_{t=0}^{T_{\text{max}}} r(s_t, a_t)] &= \mathbb{E}_{s_0\sim\mathcal{D}}[\nabla_{\theta}V^{\pi}(s_0)]\nonumber\\
&= \mathbb{E}_{s_0\sim\mathcal{D}}\left[\sum_{t=0}^{T_{\text{max}}} \sum_{a_t}\nabla_{\theta} \pi(a_t\,|\,s_t) Q^{\pi}(s_t, a_t) \right]\nonumber\\
&=  \mathbb{E}_{s_0\sim\mathcal{D}}\left[\sum_{t=0}^{T_{\text{max}}}\mathbb{E}_{a_t\sim\pi(a_t\,|\,s_t)}\left[\nabla_{\theta}\log\pi(a_t\,|\,s_t)Q^{\pi}(s_t, a_t) \right] \right]\nonumber\\
&= \mathbb{E}_{s_0\sim\mathcal{D}, \mathbf{a}\sim\pi(\cdot\,|\,s_0)}\left[\nabla_{\theta}\log\pi(\mathbf{a}\,|\,s_0)Q(\mathbf{a}\,|\,s_0) \right]\label{eq: 3-4}\\
&\quad \text{\normalfont\small\sffamily assume Q is evaluated on the full completion.}\nonumber
\end{align}

Therefore, we arrive at results consistent to where we started, approaching RLHF directly from the perspective of neural bandit structured prediction, which is a simplified version of the general RL problem where each action is selected without affecting the next state.
\section{How about we take another look at PPO’s algorithmic logic before applying it to RLHF?}
\label{sec: ppo-logic}

Many works in RLHF or LRMs use PPO \cite{schulman2017proximal} as the de facto method, incurring prohibitively high computational cost and unnecessarily complex workflows \cite{ahmadian2024back, li2023remax}. The impressive success of \say{RL-free}  methods, such as DPO \cite{rafailov2023direct}, may already suggest that RLHF is simpler than it seems. Otherwise, it would be contradictory for RL problems to be addressed effectively without relying on reinforcement learning.

Next, we revisit the principles leading to the development of PPO \cite{schulman2017proximal} to tackle a standard RL problem. In essence, PPO \cite{schulman2017proximal} is a variant of conservative policy iteration (CPI) \cite{kakade2002approximately} algorithm, implying the necessity of alternating between policy improvement and policy evaluation (This is exactly where the value function approximator, i.e., the critic, comes into play.) The following derivations are mainly borrowed from the work Trust Region Policy Optimization (TRPO) \cite{schulman2015trust}, which can be safely skipped if you are already familiar with it. A useful identity is shown in Eq. \eqref{eq: 4-1}, where the expected return of a new policy $\pi_{\text{new}}$ can be expressed in terms of advantage calculated w.r.t the old policy $\pi_{\text{old}}$. $L_{\pi_{\text{old}}}(\pi_{\text{new}})$ denotes a local approximation of $\mathcal{J}(\pi_{\text{new}})$, where states are sampled from $\rho_{\pi_{\text{old}}}(s)$ rather than $\rho_{\pi_{\text{new}}}(s)$.

\begin{align}
\mathcal{J}(\pi_{\text{new}}) &= \mathcal{J}(\pi_{\text{old}}) + \mathbb{E}_{s_0, a_0, \dots, \sim\pi_{\text{new}}} \left[\sum_{t=0}^{\infty}\gamma^t A^{\pi_{\text{old}}}(s_t, a_t) \right]\nonumber\\
&= \mathcal{J}(\pi_{\text{old}}) + \sum_s \rho_{\pi_{\text{new}}}(s) \sum_a \pi_{\text{new}}(a\,|\,s)A^{\pi_{\text{old}}}(s, a)\label{eq: 4-1}\\
L_{\pi_{\text{old}}}(\pi_{\text{new}}) &= \mathcal{J}(\pi_{\text{old}}) + \sum_s \rho_{\pi_{\text{old}}}(s) \sum_a \pi_{\text{new}}(a\,|\,s)A^{\pi_{\text{old}}}(s, a)\label{eq: 4-2}
\end{align}

The following policy improvement bound \footnote{Please refer to the original paper TRPO \cite{schulman2015trust} for a detailed proof.} introduced in TRPO is critical to guaranteeing optimizing the surrogate objective $L_{\pi_{\text{old}}}(\pi_{\text{new}})$ can lead to improvement over the true target $\mathcal{J}(\pi_{\text{new}})$.

\begin{theorem}
\label{thm: policy-improvement-bond}
Let $\alpha = D_{\text{KL}}^{\text{max}}\left( \pi_{\text{new}}, \pi_{\text{old}}\right) = \max_s D_{\text{KL}}\left( \pi_{\text{new}}, \pi_{\text{old}}\right)$\footnote{Please note that the theorem originally proposed in TRPO uses total variation divergence $D_{\text{TV}}^{\text{max}}(\pi_{\text{old}}, \pi_{\text{new}}) = \max_s D_{\text{TV}}(\pi_{\text{old}}, \pi_{\text{new}})$, which is further bounded by $D_{\text{KL}}^{\text{max}}(\pi_{\text{old}}, \pi_{\text{new}})\,\text{or}\, D_{\text{KL}}^{\text{max}}(\pi_{\text{new}}, \pi_{\text{old}})$.}, the following bound holds:

\begin{align*}
|\mathcal{J}(\pi_{\text{new}}) - L_{\pi_{\text{old}}}(\pi_{\text{new}})| \leq \frac{4\epsilon\gamma}{(1 - \gamma)^2}\alpha^2
\end{align*}
where $\epsilon=\max_{s, a}|A^{\pi_{\text{old}}}(s, a)|$
\end{theorem}

 Let $M_i(\pi) = L_{\pi_i}(\pi) - CD_{\text{KL}}^{\text{max}}(\pi, \pi_i)$, where $C=\frac{4\epsilon\gamma}{(1 - \gamma)^2}$, following the policy improvement theorem \ref{thm: policy-improvement-bond}, we can obtain:

\begin{align*}
\mathcal{J}(\pi_{i+1}) - \mathcal{J}(\pi_i) \geq M_i(\pi_{i+1}) - M_i(\pi_i)
\end{align*}

Therefore, by maximizing the surrogate (conservative) objective $M_i$, it is guaranteed that the algorithm shown in \ref{algo: 4-1} can yield a monotonically improved true policy sequence, i.e., $\mathcal{J}(\pi_0)\leq\mathcal{J}(\pi_1)\leq\mathcal{J}(\pi_2)\leq\dots$.

\IncMargin{1em}
\begin{algorithm}[H]
Initialize $\pi_0$
\BlankLine

\ForEach{$k = 0, 1, 2, \dots$, until convergence}{
	Compute all advantage values $A^{\pi_i}(s, a)$\;
	Solve the constrained optimization problem:\\
	\hspace{2em} $\pi_{i+1} = \operatorname{argmax}_{\pi} L_{\pi_i}(\pi) - CD_{\text{KL}}^{\text{max}}(\pi, \pi_i)$\;
	\hspace{2em} where $C=\frac{4\epsilon\gamma}{(1 - \gamma)^2}$,\\
	\hspace{2em} and $L_{\pi_i}(\pi) = \mathcal{J}(\pi_i) + \sum_s \rho_{\pi_i}(s)\sum_a\pi(a\,|\,s)A^{\pi_i}(s, a)$\;
}
\caption{Conservative Policy Iteration}\label{algo: 4-1}
\end{algorithm}\DecMargin{1em}

The constrained optimization problem $\pi_{i+1} = \operatorname{argmax}_{\pi} L_{\pi_i}(\pi) - CD_{\text{KL}}^{\text{max}}(\pi, \pi_i)$ is equivalent to the following expression:

\begin{align}
\max_{\theta} \mathbb{E}_{s\sim\rho_{\theta_{\text{old}}}(s), a\sim\pi_{\theta_{\text{old}}}(a\,|\,s)} \left[\frac{\pi_{\theta}(a\,|\,s)}{\pi_{\theta_{\text{old}}}(a\,|\,s)}A^{\pi_{\theta_{\text{old}}}}(s, a)  - \beta D_{\text{KL}}(\pi_{\theta}(a\,|\,s)\,\|\,\pi_{\theta_{\text{old}}}(a\,|\,s)) \right]\label{eq: 4-3}
\end{align}

In PPO \cite{schulman2017proximal}, the KL constraint is implemented implicitly through the clipping operator in tandem with a pessimistic bound (the min operator) to avoid destructively large weight updates, i.e., staying in the trust region, leading to the commonly used objective of PPO \cite{schulman2017proximal} in \eqref{eq: 4-4}.

\begin{align}
&\max_{\theta} \mathbb{E}_{s\sim\rho_{\theta_{\text{old}}}(s), a\sim\pi_{\theta_{\text{old}}}(a\,|\,s)} \left[\operatorname{min}\left(r_t(\theta)A^{\pi_{\theta_{\text{old}}}}(s, a), \operatorname{clip}\left(r_t(\theta), 1 - \epsilon, 1 + \epsilon \right)A^{\pi_{\theta_{\text{old}}}}(s, a) \right) \right]\label{eq: 4-4}\\
&\quad \text{where}\; r_t(\theta) = \frac{\pi_{\theta}(a\,|\,s)}{\pi_{\theta_{\text{old}}}(a\,|\,s)}\nonumber
\end{align}

By optimizing the surrogate objective, PPO \cite{schulman2017proximal} can achieve improved sample efficiency by performing parameter updates multiple epochs on the freshly collected samples from the last iteration (i.e., $s\sim\rho_{\theta_{\text{old}}}(s), a\sim\pi_{\theta_{\text{old}}}(a\,|\,s)$). And the critic (value network) is indispensable as the policy evaluation with value function approximators (e.g., neural nets) is realized by minimizing Bellman error (i.e., temporal difference (TD) residual). More importantly, as long as the $\max_{s, a}|A^{\pi}(s, a)|$ has not become negligible, the policy can be continuously improved.

\section{Generalized Reinforce Optimization}

In the RLHF setting where the state transition probability is deterministic, we are dealing with a much simplified optimization problem \footnote{The same unrolling trick \eqref{eq: a1-1} is applied combined with the deterministic nature of state transition probability.} \eqref{eq: 5-1}.

\begin{align}
\mathcal{J}(\pi_{\theta}) &= \mathcal{J}(\pi_{\text{old}}) + \mathbb{E}_{s_0, a_0, \dots, \sim\pi_{\theta}} \left[\sum_{t=0}^{\infty}\gamma^t A^{\pi_{\text{old}}}(s_t, a_t) \right]\nonumber\\
&= \mathcal{J}(\pi_{\text{old}}) + \mathbb{E}_{s_0 \sim \mathcal{D}} \sum_{t=0}^{T_{\text{max}}}\gamma^t \sum_{a_t} \pi_{\theta}(a_t\,|\,s_t)A^{\pi_{\text{old}}}(s_t, a_t)\label{eq: 5-1}
\end{align}

Rather than optimizing the surrogate objective $L_{\pi_{\text{old}}}(\pi_{\text{new}})$ as in the full RL setting, serving as a reliable lower bound provided $\alpha=\max_s D_{\text{KL}}\left( \pi_{\text{new}}, \pi_{\text{old}}\right)$ is reasonably small, we can directly optimizing the following objective \footnote{Please note we hereafter set $\gamma$ to $1$ as we are mostly dealing with episodic tasks in RLHF.} without the KL constraint.

\begin{align}
\max_{\theta}\mathbb{E}_{s_0 \sim \mathcal{D}} \sum_{t=0}^{T_{\text{max}}}\sum_{a_t} \pi_{\theta}(a_t\,|\,s_t)A^{\pi_{\text{old}}}(s_t, a_t)\label{eq: 5-2}
\end{align}

When the KL constraint is considered, it can be simply assimilated to the advantage function, leading to the commonly seen RLHF objective with a minor tweak where the reward has been superseded by the advantage.

\begin{align}
\max_{\theta}\mathbb{E}_{s_0 \sim \mathcal{D}} \sum_{t=0}^{T_{\text{max}}}\sum_{a_t} \pi_{\theta}(a_t\,|\,s_t)\left(A^{\pi_{\text{old}}}(s_t, a_t) - \beta \log \frac{\pi_{\theta}(a_t\,|\,s_t)}{\pi_{\theta_{\text{old}}}(a_t\,|\,s_t)} \right)\label{eq: 5-3}
\end{align}

\noindent which has a closed-form solution $\pi^* (\cdot\,|\,s_t) = \frac{1}{Z(s_t)}\pi_{\theta_{\text{old}}}(a_t\,|\,s_t)\exp(\frac{1}{\beta}A^{\pi_{\text{old}}}(s_t, a_t))$. \footnote{Please refer to the Appendix for the proof \eqref{eq: a2-1}.}. Instead of fitting it into preference models like the Bradley-Terry model, we can directly optimize the following objective:

\begin{align}
\operatorname{argmin}_{\pi_{\theta}} D_{\text{KL}}(\pi^*\,\|\,\pi_{\theta}) &= \operatorname{argmax}_{\pi_{\theta}}\mathbb{E}_{s_0 \sim \mathcal{D}} \sum_{t=0}^{T_{\text{max}}} \sum_{a_t} \pi^*(a_t\,|\,s_t)\log\pi_{\theta}(a_t\,|\,s_t)\nonumber\\
&= \operatorname{argmax}_{\pi_{\theta}}\mathbb{E}_{s_0 \sim \mathcal{D}} \sum_{t=0}^{T_{\text{max}}} \mathbb{E}_{a_t\sim \pi_{\theta_{\text{old}}}(a_t\,|\,s_t)}\left[\exp(\frac{1}{\beta}A^{\pi_{\text{old}}}(s_t, a_t))\log\pi_{\theta}(a_t\,|\,s_t)\right]\label{eq: 5-4}
\end{align}

Acute readers might have noticed that the above objective function resembles to that of advantage weighted regression (AWR) \cite{peng2019advantage} or MARWIL \cite{wang2018exponentially}, which is not a coincidence as the objective function optimized in these two works takes the following form.\footnote{Please note that when $\mathcal{D}$ (i.e., experience replay buffer) is fully static as in MARWIL, the algorithm is off-policy. While if $\mathcal{D}$ is a first-in-first-out (FIFO) queue as in AWR, the algorithm is still on-policy, resembling PPO with improved sample efficiency.}

\begin{align}
&\max_{\theta}\mathbb{E}_{s \sim \rho_{\pi_{\text{old}}}, a \sim \pi_{\theta}(a_t\,|\,s_t)}\left[A^{\pi_{\text{old}}}(s_t, a_t) - \beta \log \frac{\pi_{\theta}(a_t\,|\,s_t)}{\pi_{\theta_{\text{old}}}(a_t\,|\,s_t)} \right] \nonumber\\
&\quad = \max_{\theta}\mathbb{E}_{(s_t, a_t) \sim \mathcal{D}}\left[\exp(\frac{1}{\beta}A^{\pi_{\text{old}}}(s_t, a_t))\log\pi_{\theta}(a_t\,|\,s_t) \right] \label{eq: 5-5}
\end{align}

The objective function derived previously also share the same spirit with A-LoL\cite{baheti2023leftover}, which is motivated from off-policy gradient theorem and thus explicitly contains the problematic (sentence-level) importance weight. Intuitively, if $A^{\pi_{\text{old}}}(s_t, a_t)) > 0$, the probability of choosing the action will be pushed higher; if $A^{\pi_{\text{old}}}(s_t, a_t)) = 0$, the objective degenerates to the maximum likelihood; and if $A^{\pi_{\text{old}}}(s_t, a_t)) < 0$, the corresponding action will be discouraged but less aggressively, which can also be viewed as soft filtering of samples with strong negative advantages \footnote{It has been shown that discarding data points with negative advantages improves the learning efficiency\cite{baheti2023leftover}.}. In other words, compared to the vanilla maximum likelihood objective, advantage-weighted maximum likelihood objective equips the model with the capability of assessing the consequence of taking specific actions, mirroring how humans make decisions when facing uncertainties (i.e., Kahneman \& Tversky’s Prospect Theory \cite{tversky1992advances}), and the model can be continuously refined as long as the advantage does not vanish, revealing the inner workings of recent advances in RLHF or LRMs. The resultant algorithm can be deployed in an online fashion, mimicking PPO \cite{schulman2017proximal}, where samples are collected from the policy network in the last iteration $\pi_{\theta_{\text{old}}}$ without explicitly introducing importance weights known to introduce high variance. Furthermore, the advantageous point of not explicitly computing the importance weight $\frac{\pi_{\theta}(a\,|\,s)}{\pi_{\theta_{\text{old}}}(a\,|\,s)}$ may further unlock the potential of mixing offline language data (e.g., SFT data) with samples collected online in the fine-tuning process or exclusively employing offline data. 

A notorious problem in RLHF is maximizing rewards solely would lead to degenerate models due to insufficient exploration or reward hacking \footnote{Please refer to Tomek Korbak's blog \href{https://tomekkorbak.com/2022/05/20/rl-with-kl-penalties-bayesian-inference/}{RL with KL penalties is better viewed as Bayesian inference} for further insights.}, i.e., distribution collapse where the generation diversity has been compromised for eliciting desirable behaviour steered by certain rewards, resulting in using KL divergence to penalize the policy network deviating substantially from the reference network. The recent advances in LRMs have shown that it may be not necessary to incorporate KL regularization into the reward maximization objective, partly attributed to rule-based reward functions which can reduce the risk of reward hacking, thus promoting sufficient exploration. An oracle reward model is expected to sort candidate responses in a descending order proportional to the degree to which they match the task criteria, i.e., a bijective mapping from the space of all possible candidates to monotonic numerical ratings. In other words, a flawed reward model is prone to induce candidate solutions inseparable to each other, i.e., diminished generation diversity. Even though the term $\exp(\frac{1}{\beta}A^{\pi_{\text{old}}}(s_t, a_t))$ can mitigate distribution collapse to certain degree as it naturally contains the maximum likelihood as a special case, we hypothesize it is still desirable to incorporate mechanisms in order to explicitly encourage separation between candidate responses. Motivated by the success of RL-free methods, we hypothesize it is promising to add the contrastive ingredient to strengthen advantage-weighted maximum likelihood.

We are in a position to introduce the Generalized Reinforce Optimization (GRO), unifying RL-based and RL-free pathways in RLHF.

\begin{align}
J_{\text{GRO}}(\pi_{\theta}) = \mathbb{E}_{s_0 \sim \mathcal{D}} \sum_{t=0}^{T_{\text{max}}} \mathbb{E}_{a_t\sim \pi_{\theta_{\text{old}}}(a_t\,|\,s_t)}\left[\omega(\alpha(\log(\pi_{\theta}^{\text{sg}}(a_t\,|\,s_t)) - \varepsilon_{\ast}))\upsilon(\frac{1}{\beta}A^{\pi_{\text{old}}}(s_{T_{\text{max}}}))\log\pi_{\theta}(a_t\,|\,s_t)\right]\label{eq: 5-6}
\end{align}

\noindent where $\omega(\cdot)$ takes into account the difference between the log likelihood of the predicted action $\pi_{\theta}^{\text{sg}}(a_t\,|\,s_t)$ ($\operatorname{sg}$ means stop gradient) and an anchor value $\varepsilon_{\ast}$ to encourage separation while preventing  excessively pushing away those well-distinguished samples, $A^{\pi_{\text{old}}}(s_{T_{\text{max}}}) = R(s_{T_{\text{max}}}) - B$ is the sequence-level advantage, and $\upsilon(\cdot)$ is a monotonically increasing function. Although the equation in \eqref{eq: 5-4} indicates that token-level advantage is theoretically feasible, we adhere to the setting of neural bandit structured prediction where a full completion is regarded as the single action.  If we return to the language modelling notation, the above objective can be rewritten as follows.

\begin{align}
J_{\text{GRO}}(\pi_{\theta}) &= \mathbb{E}_{p(\mathbf{x})\sim\mathcal{D}}  \sum_{t=0}^{T_{\text{max}}} \mathbb{E}_{a_t\sim \pi_{\theta_{\text{old}}}(y_t\,|\,\mathbf{y}_{<t}, \mathbf{x})} \left[\omega(\alpha(\log\pi_{\theta}^{\text{sg}}(y_t\,|\,\mathbf{y}_{<t}, \mathbf{x}) - \varepsilon_{\ast}))\right.\nonumber\\
&\quad \left. \upsilon(\frac{1}{\beta}A^{\pi_{\text{old}}}(\mathbf{y}, \mathbf{x})) \log\pi_{\theta}(y_t\,|\,\mathbf{y}_{<t}, \mathbf{x})\right]\label{eq: 5-7}\\
\hat{g} &= \frac{1}{N}\sum_{i=1}^N\sum_{t=0}^{T_{\text{max}}} \omega(\alpha(\log\pi_{\theta}^{\text{sg}}(y_t^i\,|\,\mathbf{y}^i_{<t}, \mathbf{x}) - \varepsilon_{\ast}))\upsilon(\frac{1}{\beta}A^{\pi_{\text{old}}}(\mathbf{y}^i, \mathbf{x})) \nabla_{\theta}\log\pi_{\theta}(y_t^i\,|\,\mathbf{y}_{<t}^i, \mathbf{x})\label{eq: 5-8}
\end{align}

A wide variety of RL-based or RL-free methods, at least those methods introduced in this article, can be subsumed under the proposed GRO objective, enhancing the previous advantage-weighted maximum likelihood with a dynamic weighting function that gauges the distance between the predicted sequence and dynamically selected anchor sequences in terms of log likelihood \footnote{It it worth noting that calculating the distance (esp., at a sentence-level) in terms of log probabilities is not computationally light for language modelling tasks, considering the $\operatorname{logsumexp}$ over the size of large vocabulary. But we leave exploring potential metric functions to the future work.} to ensure sufficient separation in generated completions (i.e., counteracting the potential of distribution collapse). Specifically, in RL-based algorithms, e.g., RLOO \cite{ahmadian2024back}, ReMax \cite{li2023remax}, $\omega(\cdot) = 1$, and $\upsilon(\cdot)$ is implemented as an identity function. In RL-free methods, e.g., DPO \cite{rafailov2023direct}, KTO \cite{ethayarajh2024kto}, CPL \cite{hejna2023contrastive}, the design emphasis has been focused on $\omega(\cdot)$ with distances measured by log ratios w.r.t a base model $\pi_{\text{ref}}$, failing to tap the potential of the advantage function as evidenced by using raw rewards ($+1/-1$ for positive/negative responses). It is worth noting that RL-based methods adapted from PPO \cite{schulman2017proximal}, e.g., GRPO \cite{shao2024deepseekmath} and REINFORCE++ \cite{hu2025reinforce++}, share a higher similarity with the proposed GRO objective with a hard clipping operator mimicking the dynamic weighting function $\omega(\cdot)$, shedding light on the impressive performance achieved by these methods. We present a comparison of recent advances in RL-based and RL-free approaches in RLHF under the proposed unified framework of GRO in Tab. \ref{tab: 5-1}, bridging the gap between two ostensibly dichotomous lines of research.

\begin{sidewaystable}[!htb]
	\renewcommand{\arraystretch}{1.1}
	\setlength{\tabcolsep}{3.0pt}
	\centering
	\caption{Comparison of a set of RL-based and RL-free methods in RLHF via the lens of GPO}
	\label{tab: 5-1}
	\begin{tabular}{cc|c|c|c|c}
		\toprule
		&& $\omega(\cdot)$ & $\varepsilon_{\ast}$ & $B$ & $A(\cdot)$ \\
		\midrule
		RLOO&& 1 & n/a & $\frac{1}{N-1}\sum_{j\neq i}R(\mathbf{y}^j)$ & $R(\mathbf{y}^i) - \frac{1}{N-1}\sum_{j\neq i}R(\mathbf{y}^j)$\\
		ReMax&& 1 & n/a & $R(\mathbf{y})^{\text{greedy}}$ & $R(\mathbf{y}^i) - R(\mathbf{y})^{\text{greedy}}$ \\
		\multirow{2}{*}{\centering GRPO}&& \multirow{2}{*}{\centering $\operatorname{clip}(\cdot)$} & \multirow{2}{*}{\centering $\{1 - \epsilon, 1 + \epsilon \}$} & \multirow{2}{*}{\centering $\mu_g = \sum_{i=1}^N R(\mathbf{y}^i)$} & $\frac{R(\mathbf{y}^i) - \mu_g}{\sigma_g}$\\
		&& & & & where $\sigma_g = \sqrt{\frac{\sum_{i=1}^N (R(\mathbf{y}^i) - \mu_g)^2}{N}}$\\
		\multirow{2}{*}{\centering REINFORCE++}&& \multirow{2}{*}{\centering$\operatorname{clip}(\cdot)$} & \multirow{2}{*}{\centering$\{1 - \epsilon, 1 + \epsilon \}$} & \multirow{2}{*}{\centering$\mu_g = \sum_{i=1}^N R(\mathbf{y}^i)$} & $\gamma^{T_{\text{max}}-t}(R(\mathbf{y}^i) - \mu_g)$\\
		&& & & & $+\sum_{k=t}^{T_{\text{max}}} \gamma^{k-t} \log\frac{\pi_{\theta}(y_t\,|\,\mathbf{y}_{<t}^i, \mathbf{x})}{\pi_{\text{ref}(y_t\,|\,\mathbf{y}_{<t}^i, \mathbf{x})}}$\\
		DPO &&$1 - \sigma(\cdot)$ &  $\log\frac{\pi_{\theta}(\mathbf{y}^{\neg\operatorname{sign}(i)}\,|\,\mathbf{x})}{\pi_{\text{ref}}(\mathbf{y}^{\neg\operatorname{sign}(i)}\,|\,\mathbf{x})}$ & 0 & $R(\mathbf{y}^i)$\\
		\multirow{2}{*}{\centering KTO} && \multirow{2}{*}{\centering$\sigma(\cdot)(1 - \sigma(\cdot))$}& $\max(0, \frac{1}{m}\sum_i \log\frac{\pi_{\theta}(\mathbf{y}^{j}\,|\,\mathbf{x}^i)}{\pi_{\text{ref}}(\mathbf{y}^{j}\,|\,\mathbf{x}^i)})$ & \multirow{2}{*}{\centering 0} & \multirow{2}{*}{\centering$R(\mathbf{y}^i)$}\\
		&& & $ j = (i + 1) \operatorname{mod} m$ & \\
		CPL && $1 - \sigma(\cdot)$ & $\log\frac{\pi_{\theta}(\mathbf{y}^{\neg\operatorname{sign}(i)}\,|\,\mathbf{x})}{\pi_{\text{ref}}(\mathbf{y}^{\neg\operatorname{sign}(i)}\,|\,\mathbf{x})}$ & 0 & $R(\mathbf{y}^i)$\\
		\bottomrule
	\end{tabular}
\end{sidewaystable}
\section{Concluding Thoughts}
To wrap up, this article has explored recent advancements in RL-based and RL-free methods for RLHF, uncovering that the REINFORCE-style gradient estimator serves as the central mechanism stimulating exploration with evaluative feedback. We then have revisited core RL principles and the algorithmic foundation of PPO \cite{schulman2017proximal}, recognizing that the deterministic state transitions in RLHF can simplify the optimization process. Finally, we have introduced GRO, a unified learning framework for RLHF that encompasses various RL-based and RL-free approaches. We hope this article aids researchers and practitioners in fields such as RLHF and LLMs in deepening their understanding of core mechanisms, and we eagerly anticipate the community’s efforts to empirically validate GRO while welcoming constructive feedback.
\section*{Acknowledgements}
We express our gratitude to \textit{Mathematical Foundations of Reinforcement Learning} \cite{zhao2024mathematical} by \href{https://www.shiyuzhao.net/}{Prof. Shiyu Zhao}, a resource that has proven invaluable in navigating the complexities of reinforcement learning (RL). Its detailed mathematical derivations have significantly enriched our understanding of the fundamental principles underlying general RL algorithms, rendering it a highly recommended read for those seeking to overcome the high entry barrier of RL.

We also extend our appreciation to Grok, developed by \href{https://x.ai/}{xAI}, for its assistance in refining some expressions of this article.

\bibliographystyle{splncs04}	
\bibliography{reference}

\begin{thebibliography}{10}
\providecommand{\url}[1]{\texttt{#1}}
\providecommand{\urlprefix}{URL }
\providecommand{\doi}[1]{https://doi.org/#1}

\bibitem{ahmadian2024back}
Ahmadian, A., Cremer, C., Gall{\'e}, M., Fadaee, M., Kreutzer, J., Pietquin,
  O., {\"U}st{\"u}n, A., Hooker, S.: Back to basics: Revisiting reinforce style
  optimization for learning from human feedback in llms. arXiv preprint
  arXiv:2402.14740  (2024)

\bibitem{baheti2023leftover}
Baheti, A., Lu, X., Brahman, F., Bras, R.L., Sap, M., Riedl, M.: Leftover
  lunch: advantage-based offline reinforcement learning for language models.
  arXiv preprint arXiv:2305.14718  (2023)

\bibitem{ethayarajh2024kto}
Ethayarajh, K., Xu, W., Muennighoff, N., Jurafsky, D., Kiela, D.: Kto: Model
  alignment as prospect theoretic optimization. arXiv preprint arXiv:2402.01306
   (2024)

\bibitem{gao2022simulating}
Gao, G., Choi, E., Artzi, Y.: Simulating bandit learning from user feedback for
  extractive question answering. arXiv preprint arXiv:2203.10079  (2022)

\bibitem{hejna2023contrastive}
Hejna, J., Rafailov, R., Sikchi, H., Finn, C., Niekum, S., Knox, W.B., Sadigh,
  D.: Contrastive preference learning: learning from human feedback without rl.
  arXiv preprint arXiv:2310.13639  (2023)

\bibitem{hu2025reinforce++}
Hu, J.: Reinforce++: A simple and efficient approach for aligning large
  language models. arXiv preprint arXiv:2501.03262  (2025)

\bibitem{kakade2002approximately}
Kakade, S., Langford, J.: Approximately optimal approximate reinforcement
  learning. In: Proceedings of the nineteenth international conference on
  machine learning. pp. 267--274 (2002)

\bibitem{kreutzer2017bandit}
Kreutzer, J., Sokolov, A., Riezler, S.: Bandit structured prediction for neural
  sequence-to-sequence learning. arXiv preprint arXiv:1704.06497  (2017)

\bibitem{lam2018reinforcement}
Lam, T.K., Kreutzer, J., Riezler, S.: A reinforcement learning approach to
  interactive-predictive neural machine translation. arXiv preprint
  arXiv:1805.01553  (2018)

\bibitem{li2023remax}
Li, Z., Xu, T., Zhang, Y., Lin, Z., Yu, Y., Sun, R., Luo, Z.Q.: Remax: A
  simple, effective, and efficient reinforcement learning method for aligning
  large language models. arXiv preprint arXiv:2310.10505  (2023)

\bibitem{nguyen2017reinforcement}
Nguyen, K., Daum{\'e}~III, H., Boyd-Graber, J.: Reinforcement learning for
  bandit neural machine translation with simulated human feedback. arXiv
  preprint arXiv:1707.07402  (2017)

\bibitem{peng2019advantage}
Peng, X.B., Kumar, A., Zhang, G., Levine, S.: Advantage-weighted regression:
  Simple and scalable off-policy reinforcement learning. arXiv preprint
  arXiv:1910.00177  (2019)

\bibitem{rafailov2023direct}
Rafailov, R., Sharma, A., Mitchell, E., Manning, C.D., Ermon, S., Finn, C.:
  Direct preference optimization: Your language model is secretly a reward
  model. Advances in Neural Information Processing Systems  \textbf{36},
  53728--53741 (2023)

\bibitem{schulman2015trust}
Schulman, J., Levine, S., Abbeel, P., Jordan, M., Moritz, P.: Trust region
  policy optimization. In: International conference on machine learning. pp.
  1889--1897. PMLR (2015)

\bibitem{schulman2017proximal}
Schulman, J., Wolski, F., Dhariwal, P., Radford, A., Klimov, O.: Proximal
  policy optimization algorithms. arXiv preprint arXiv:1707.06347  (2017)

\bibitem{shao2024deepseekmath}
Shao, Z., Wang, P., Zhu, Q., Xu, R., Song, J., Bi, X., Zhang, H., Zhang, M.,
  Li, Y., Wu, Y., et~al.: Deepseekmath: Pushing the limits of mathematical
  reasoning in open language models. arXiv preprint arXiv:2402.03300  (2024)

\bibitem{tversky1992advances}
Tversky, A., Kahneman, D.: Advances in prospect theory: Cumulative
  representation of uncertainty. Journal of Risk and uncertainty  \textbf{5},
  297--323 (1992)

\bibitem{wang2018exponentially}
Wang, Q., Xiong, J., Han, L., Liu, H., Zhang, T., et~al.: Exponentially
  weighted imitation learning for batched historical data. Advances in Neural
  Information Processing Systems  \textbf{31} (2018)

\bibitem{xie2025logic}
Xie, T., Gao, Z., Ren, Q., Luo, H., Hong, Y., Dai, B., Zhou, J., Qiu, K., Wu,
  Z., Luo, C.: Logic-rl: Unleashing llm reasoning with rule-based reinforcement
  learning. arXiv preprint arXiv:2502.14768  (2025)

\bibitem{zhao2024mathematical}
Zhao, S.: Mathematical foundations of reinforcement learning. Mathematical
  Foundation of Reinforcement Learning p.~195 (2024)

\bibitem{ziegler1909fine}
Ziegler, D.M., Stiennon, N., Wu, J., Brown, T.B., Radford, A., Amodei, D.,
  Christiano, P., Irving, G.: Fine-tuning language models from human
  preferences, 2020. URL https://arxiv. org/abs p.~14 (1909)

\end{thebibliography}
\clearpage

\begin{appendices}
\appendix
\renewcommand{\thesection}{\Alph{section}.\arabic{section}}
\setcounter{section}{0}
\counterwithin{figure}{section}
\counterwithin{table}{section}
\renewcommand{\theequation}{A.\arabic{equation}}
\setcounter{equation}{0}

\section{Proof of $\nabla_{\theta}V^{\pi}(s)$}

The following proof uses the unrolling trick. We encourage interested readers to consult the book \textit{Mathematical Foundations of Reinforcement Learning} \cite{zhao2024mathematical} for a mathematically elegant proof of the Policy Gradient Theorem, featuring matrix equations and the Kronecker product, as well as an equivalent formulation based on the average reward metric.

\begin{align}
\nabla V^{\pi}(s) &= \nabla \left[\sum_a \pi(a\,|\,s)Q^{\pi}(s, a) \right]\nonumber\\
	&= \sum_a\left[\nabla \pi(a\,|\,s) Q^{\pi}(s, a)  + \pi(a\,|\,s) \nabla Q^{\pi}(s, a) \right]\nonumber\\
	&= \sum_a\left[\nabla \pi(a\,|\,s) Q^{\pi}(s, a)  + \pi(a\,|\,s) \nabla (r(s, a) + \gamma\sum_{s^{\prime}}P(s^{\prime}\,|\,s, a)V^{\pi}(s^{\prime}))\right]\nonumber\\
	&=\sum_a\left[\nabla \pi(a\,|\,s) Q^{\pi}(s, a)  + \pi(a\,|\,s) \gamma\sum_{s^{\prime}}P(s^{\prime}\,|\,s, a)\nabla V^{\pi}(s^{\prime})\right]\nonumber\\
	&= \sum_a\left[\nabla \pi(a\,|\,s) Q^{\pi}(s, a)  + \pi(a\,|\,s) \gamma\sum_{s^{\prime}}P(s^{\prime}\,|\,s, a) \right. \nonumber\\
	&\quad\quad \left. \sum_{a^{\prime}} [ \nabla \pi(a^{\prime}\,|\,s^{\prime}) Q^{\pi}(s^{\prime}, a^{\prime}) + \pi(a^{\prime}\,|\,s^{\prime}) \gamma\sum_{s^{\prime\prime}}P(s^{\prime\prime}\,|\,s^{\prime}, a^{\prime})\nabla V^{\pi}(s^{\prime\prime}) ] \right]\quad \text{(unrolling)}\nonumber\\
	&= \sum_a\nabla \pi(a\,|\,s) Q^{\pi}(s, a) + \gamma \sum_a \pi(a\,|\,s)\sum_{s^{\prime}}P(s^{\prime}\,|\,s, a) \sum_{a^{\prime}} \nabla \pi(a^{\prime}\,|\,s^{\prime}) Q^{\pi}(s^{\prime}, a^{\prime}) \nonumber\\
	&\quad\quad + \gamma^2 \sum_a \pi(a\,|\,s)\sum_{s^{\prime}}P(s^{\prime}\,|\,s, a)\sum_{a^{\prime}}\pi(a^{\prime}\,|\,s^{\prime})\sum_{s^{\prime\prime}}P(s^{\prime\prime}\,|\,s^{\prime}, a^{\prime}) \nabla V^{\pi}(s^{\prime\prime})\nonumber\\
	&= \sum_a\nabla \pi(a\,|\,s) Q^{\pi}(s, a) + \gamma \sum_{s^{\prime}}P(s^{\prime}\,|\,s) \sum_{a^{\prime}} \nabla \pi(a^{\prime}\,|\,s^{\prime}) Q^{\pi}(s^{\prime}, a^{\prime}) \nonumber \\
&\quad\quad + \gamma^2 \sum_{s^{\prime}}P(s^{\prime}\,|\,s) \sum_{s^{\prime\prime}}P(s^{\prime\prime}\,|\,s^{\prime}) \nabla V^{\pi}(s^{\prime\prime})\nonumber\\
	&=\sum_{s^{\prime}}\underbrace{\sum_{k=0}^{\infty} \gamma^k \operatorname{Pr}_{\pi}(S_{t_k}=s^{\prime}\,|\,S_{t_1}=s)}_{\operatorname{Pr}_{\pi}(s^{\prime}\,|\,s)} \sum_{a} \nabla \pi(a\,|\,s^{\prime})Q^{\pi}(s^{\prime}, a) \label{eq: a1-1}
\end{align}
\clearpage

\section{Proof the closed form solution of the objective in \eqref{eq: 5-3}}

The following proof is similar to the derivation in DPO \cite{rafailov2023direct}.

\begin{align}
&\max_{\pi_{\theta}} \sum_a \pi_{\theta}(a\,|\,s) \left(A^{\pi_{\text{old}}}(s, a) - \beta \log \frac{\pi_{\theta}(a\,|\,s)}{\pi_{\theta_{\text{old}}}(a\,|\,s)}\right) \nonumber\\
&\quad\quad = \min_{\pi_{\theta}} \sum_a \pi_{\theta}(a\,|\,s) \left(\log \frac{\pi_{\theta}(a\,|\,s)}{\pi_{\theta_{\text{old}}}(a\,|\,s)} - \frac{1}{\beta}A^{\pi_{\text{old}}}(s, a) \right)\nonumber\\
&\quad\quad = \min_{\pi_{\theta}} \sum_a \pi_{\theta}(a\,|\,s)\left(\log\frac{\pi_{\theta}(a\,|\,s)}{\frac{1}{Z(s)}\pi_{\theta_{\text{old}}}(a\,|\,s)\exp(\frac{1}{\beta}A^{\pi_{\text{old}}}(s, a))} - \log Z(s) \right)\nonumber\\
&\quad\quad = \min_{\pi_{\theta}} \mathrm{D}_{\text{KL}}(\pi_{\theta}(a\,|\,s)\,\|\, \pi^*(a\,|\,s))\; \text{where}\; \pi^*(a\,|\,s) = \frac{1}{Z(s)}\pi_{\theta_{\text{old}}}(a\,|\,s)\exp(\frac{1}{\beta}A^{\pi_{\text{old}}}(s, a)) \label{eq: a2-1}
\end{align}

\end{appendices}

\end{document}